\documentclass[twoside,twocolumn,10pt]{article}

\usepackage{wscg}           
\usepackage{calc}
\usepackage{amssymb}
\usepackage{amstext}
\usepackage{amsmath}
\usepackage{amsthm}
\usepackage{multicol}
\usepackage{pslatex}
\usepackage{mathtools}
\DeclarePairedDelimiter{\floor}{\lfloor}{\rfloor}

\RequirePackage{ifpdf}
\ifpdf
 \RequirePackage[pdftex]{graphicx}
 \RequirePackage[pdftex]{color}
\else
 \RequirePackage[dvips,draft]{graphicx}
 \RequirePackage[dvips]{color}
\fi

\usepackage{nopageno}       

\title{A framework for robust object multi-detection with a vote aggregation and a cascade filtering}

\author{
\parbox{0.30\textwidth}{\centering
Grzegorz Kurzejamski\\[1mm]
Institute of Microelectronics and Optoelectronics\\
Warsaw University of Technology\\
00-661 Warsaw, Poland\\[1mm]
Lingaro Sp. z o.o.\\
Pu\l awska 99a\\
02-595 Warsaw, Poland\\[1mm]
grzegorz.kurzejamski@gmail.com
}
\hspace{0.05\textwidth}
\parbox{0.30\textwidth}{\centering
Jacek Zawistowski\\[1mm]
Institute of Microelectronics and Optoelectronics\\
Warsaw University of Technology\\
00-661 Warsaw, Poland\\[1mm]
Lingaro Sp. z o.o.\\
Pu\l awska 99a\\
02-595 Warsaw, Poland\\[1mm]
jzawisto@gmail.com
}
\hspace{0.05\textwidth}
\parbox{0.30\textwidth}{\centering
Grzegorz Sarwas\\[1mm]
Lingaro Sp. z o.o.\\
Pu\l awska 99a\\
02-595 Warsaw, Poland\\[1mm]
grzegorz.sarwas@gmail.com
}
}

\usepackage{url}
\makeatletter
\def\Uslash{\mathbin{\mathchar`\/}\@ifnextchar{/}{\kern-.15em}{}}
\g@addto@macro\UrlSpecials{\do \/ {\Uslash}}
\def\Ucolon{\mathbin{\mathchar`:}\@ifnextchar{/}{\kern-.1em}{}}
\g@addto@macro\UrlSpecials{\do : {\Ucolon}}
\makeatother

\begin{document}

\twocolumn[{\csname @twocolumnfalse\endcsname

\maketitle  

\begin{abstract}
\noindent
This paper presents a framework designed for the multi-object detection purposes and adjusted for the application of product search on the market shelves. The framework uses a single feedback loop and a pattern resizing mechanism to demonstrate the top effectiveness of the state-of-the-art local features. A high detection rate with a low false detection chance can be achieved with use of only one pattern per object and no manual parameters adjustments. The method incorporates well known local features and a basic matching process to create a reliable voting space. Further steps comprise of metric transformations, graphical vote space representation, two-phase vote aggregation process and a cascade of verifying filters. 

\end{abstract}

\subsection*{Keywords}
Computer Vision, Image Analysis, Multiple Object Detection, Object Localization, Pattern Matching.

\vspace*{1.0\baselineskip}
}]


\section{Introduction}

As computer vision algorithms are being vastly developed in many fields, it is still very unlikely to create production-class detection systems for various applications. This paper is focused on the problem of detection of retail products shown on the market shelves and displays. This particular application demands usage of a multi-object multi-detection system (possible many instances of the different object classes in one scene). The patterns in this case are generic graphics most of the time and geometric transformations in the scene are much simpler than those found in natural scenery. Even though, it's still a demanding task as brands' numbers are counted in hundreds and each brand can have up to a thousand different wrapping layouts. Moreover, each brand has some percentage of common graphics present, for example logos. There are no standards in size or shape of the products. It's very expensive to take dozens of photos of each sample wrapping in different environmental conditions as well, so learning methods may be inefficient in real applications.

There are many approaches to multi-detection systems with generic graphics as patterns. The most common is the local features approach, where system operates on descriptors containing information about a locality of a particular graphical element. Local features give many possibilities for optimization for multi-pattern databases. In the application of retail product search we assumed that the detection rate, sufficient localization precision and low false detection rate are of the most importance. Computational efficiency is on the second place, as we do not assume real-time processing.

A multi-object detection system has to have a localization step, that may be used to divide the approaches into several groups. The first group may be a general object detection approach, which contain a saliency detector and a contextual image clustering. These methods are independent of any pattern and try to differ the background from foreground objects. There are some visual features, as edges and a frequency response, that can show areas of the image, that can be taken as an object. Another example of a general clustering approach has been presented in the work of Iwanowski \emph{et al.} \cite{Polki01}. Unfortunately, this particular method fails in many scenes, as it needs very explicit shelves' and products' edges visible. Generic saliency methods failed in every one of the test photos, as scenes with products on shelves are salient in almost every spot. The second group of localization methods may use a voting scheme and local features. Local features in the scene can be matched against local features in the pattern. Consequent correspondences can be used to localize an object of a particular type in the scene. The complexity of such search can be minimized by using multiple detection stages, starting from the general search (logo or brand search) to a specialized identification (identification of the brand's member). 

The last group of the localization approaches uses dense feature matching against a whole pattern database for each possible window in the scene. Algorithm has to generate a set of windows in any position and of any size. Such approach, called usually the sliding window approach, has been vastly used for object search purposes. Each window has to be processed as a standalone image in search for one instance of the object. It's obvious, that majority of the generated windows will not fit perfectly into object's envelope. The number of windows can be counted in thousands even in optimized window search. Each window has to be analyzed by a global image descriptor or a set of local descriptors. These descriptors have to be matched against the whole pattern database. The most advanced methods use hashing to minimize computational complexity in case of a large pattern database. Bag of words approach gives good results for local features as well. Despite of many optimizations in window search algorithms, such approach can be still too complex for modern machines in case of the analyzed applications. On the other hand, there are known well optimized multi-class multi-detection systems using modified HOG descriptors and LSH hashing methods. Unfortunately such methods use learning process and are not suitable for detecting specific, generic graphics with high amount of common visual elements. Many systems use global similarity metric, that gives good results in case of KNN (k nearest neighbours) queries. It's important though to create highly robust filter, that rejects false detections, as KNN queries don't provide information whether the best result can be accepted as a match. Simple distance thresholding may be not sufficient to accomplish this task effectively.

This paper presents the multi-detection system based on a method of vote space analysis. System is based on the invention shown in~\cite{Pse01}. System uses local features and voting mechanism for localization and a cascade of filters to reject false detections presented in~\cite{Pse02}. System is ready to use for a multiple stage detection and has linear scalability in regards to the pattern number. Using simple parameter automation mechanisms allowed maximization of detection rate. Achieving high amount of control over false detection response was the most important aspect of system's application. We used implementation of SIFT algorithm for tests, but presented approach can be used with any feature points containing scale and rotation information.

\section{Related work}

\noindent There are multiple works presenting building of a vote space for multi-detection purposes. Lowe in~\cite{Sift02} proposes generalized Hough Transform for clustering the vote space. Authors of \cite{AzadAD09a} create a 4D voting space and use combination of Hough, RANSAC and Least Squares Homography Estimation in order to detect and accept potential objects' instances. Zickler in \emph{et al.}~\cite{ZicklerE07} use angle differences criterion in addition to RANSAC mechanism and a vote number threshold. Zickler \emph{et al.} in \cite{ZicklerV06} use a custom probabilistic model in addition to the Hough algorithm. Branch-and-bound approaches as in~\cite{Yeh09} are promising for multi-detection purposes in conjunction with Bag-of-words descriptors. Viola and Jones in~\cite{Haar01} developed cascade of boosted features, that can efficiently detect multiple instances of the same object in one pass of the detection process. The method needs a time consuming, learning process on thousands of images. Method has been tested mostly on general objects, as people, cars, faces. Blaschko and Lampert in~\cite{BlaschkoL08} use SVM to enhance sliding window process. Efficient subwindows search has been used in~\cite{LampertBH08}. A most straightforward method of multi-detection is using all of the windows from sliding window algorithm, as used in Sarwas' and Skoneczny's work~\cite{VarFil01}. High effectiveness can be achieved with Histogram of Oriented Gradients~\cite{Hog01} and Deformable Part Models~\cite{Dpm01}. Interesting use of DPM and LSH can be found in the work of Dean \emph{et al.} \cite{GoogleDetection01}. The biggest drawback of the Deformable Part Models and Histogram of Oriented Gradients for analyzed application is that they usually need learning stage and are not rotation invariant.

\section{System overview}

\begin{figure*}[!htb]
\begin{minipage}[!htb]{.3\linewidth}
  \centering
  \centerline{\includegraphics[width=5cm]{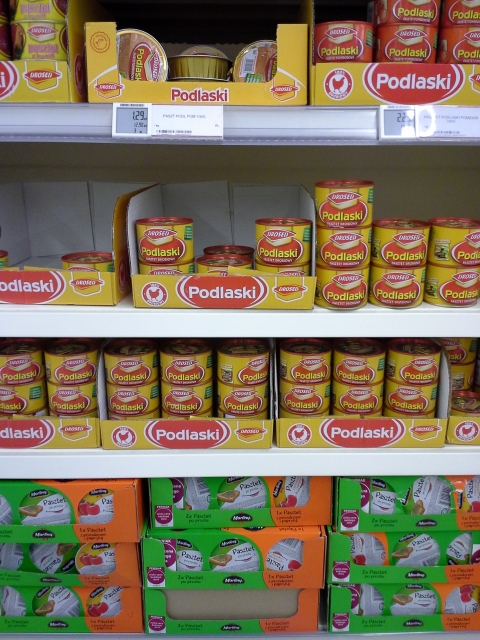}}
  \centerline{(a) Scene image.}\medskip
  \label{fig:votemap_obj1}
\end{minipage}
\hfill
\begin{minipage}[!htb]{.3\linewidth}
  \centering
  \centerline{\includegraphics[width=5cm]{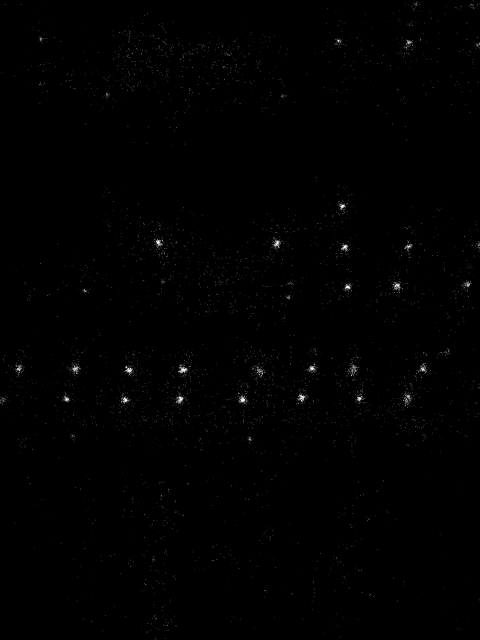}}
  \centerline{(b) Vote image.}\medskip
  \label{fig:votemap_obj2}
\end{minipage}
\hfill
\begin{minipage}[!htb]{.3\linewidth}
  \centering
  \centerline{\includegraphics[width=5cm]{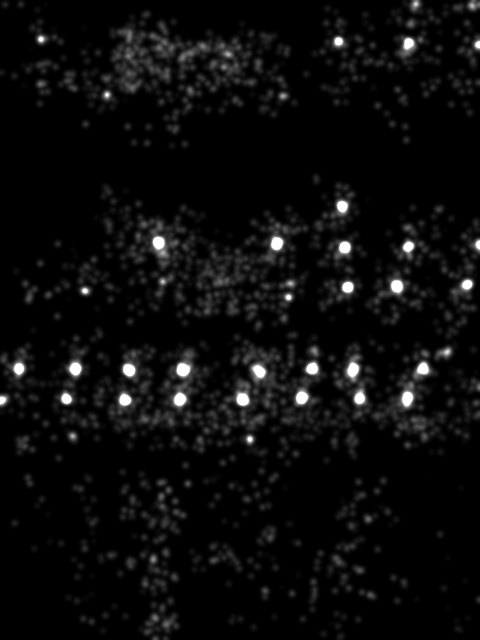}}
  \centerline{(c) Blurred and normalized vote image.}\medskip
  \label{fig:votemap_obj3}
\end{minipage}
\caption{Sample of a vote image generated while localizing a \emph{Drosed} product.}
\label{fig:votemap}
\end{figure*}

\noindent Presented system uses scale and rotation invariant local features for object detection. The core of the system is voting schema connected with a cascade of filters. Given a particular pattern we create the cascade of resized patterns. We extract local features in both the scene and the pattern images. Features from the two groups are matched against each other with a FLANN \cite{FLANN01} algorithm. Correspondences are filtered with a contrast data and a color distance criteria. The threshold value for the contrast data distance is calculated as a middle value between the lowest and the highest distance values found in correspondence set. Color distance thresholding function does not apply for some values of HSL channels of a matched feature points pair. The contrast data distance is transformed to create the Adjacency value with a function: 
\begin{equation}
 \mbox{\textit{adj}}(m) = 1 - \left(
 \frac{dist(m)}{thr}
  \right)^2,
\end{equation}
 where \emph{m} denotes the feature points match, \emph{dist(m)} denotes distance between feature points in match \emph{m} and \emph{thr} is a distance threshold value.

Each correspondence is used as a vote in a multi-dimensional vote space. The vote space is not analyzed in a direct manner. It is projected onto the X, Y plane, where X and Y dimensions are identical to X and Y dimensions of the scene image. The adjacency values of each vote are summed for each (x, y) bucket and used as a cue to create a single channel image (called a vote image in this paper) of the same size as the scene. Adjacency values in the vote image can be normalized, and the image can be blurred to make it possible for human to analyze it and evaluate the efficiency of matching process. Such blurred and normalized vote image can be seen in Figure~\ref{fig:votemap}. Vote image is processed by a graphical local maxima detector. We found that Good Features To Track \cite{Goodfeatures01} works very well for this task. Local maxima in the vote image are further called propositions. Propositions are sorted by adjacency sum value in descending order. Each proposition is a center of a potential object instance in the image.

\begin{figure}[!htb]
\begin{minipage}[!htb]{\linewidth}
  \centering
  \centerline{\includegraphics[width=7.5cm]{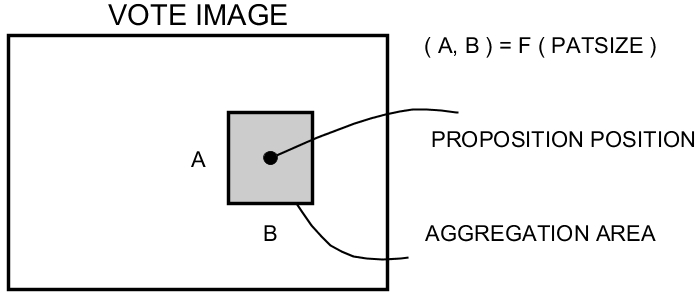}}
  \centerline{Pass 1 of the aggregation.}\medskip
  \label{fig:aggreg_1}
\end{minipage}
  
\begin{minipage}[!htb]{\linewidth}
  \centering
  \centerline{\includegraphics[width=7.5cm]{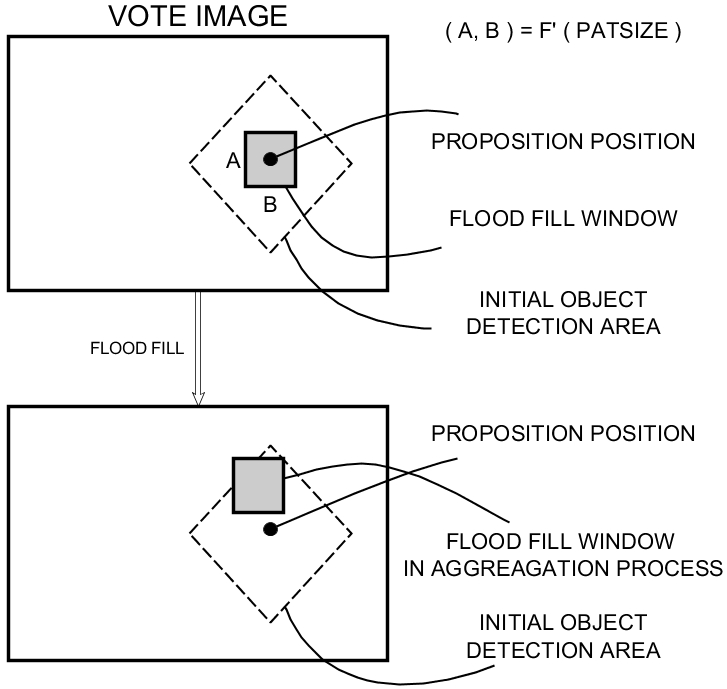}}
  \centerline{Pass 2 of the aggregation.}\medskip
  \label{fig:aggreg_2}
\end{minipage}

\caption{Aggregation process with aggregation window.}
\label{fig:aggregation_schema}
\end{figure}

For each proposition, starting from the one with the highest adjacency sum, we perform a vote aggregation and a cascade filtering. Each of the filters in a cascade can accept or reject current vote aggregation process. Any rejection will lead to dropping the aggregation process and removing the processed proposition from the propositions sorted queue. Vote aggregation is a two-pass algorithm. Pass one of the aggregation collects all of the votes in a local area of proposition's position. After gathering of all of the votes in the local area, the unique filtering (discrabed later in this paper) is performed and the resulting group of votes is tested against a cascade of filters. In the second pass of the process the aggregation is conducted with the Flood Fill algorithm, starting from the proposition's position. The Flood Fill range is constrained by a scaled down object's envelope. Sizes of the local area in the pass one and of the Flood Fill search window in the pass two rely on a pattern size. The idea is presented in Figure~\ref{fig:aggregation_schema}. In pass two we've already got the estimation of the the object's envelope after analysis of the votes' data from the first pass. Second pass of the algorithm contains unique filtering and cascade filtering as well.

The unique filtering takes place after the vote aggregation and before the cascade filtering in each pass. It is a simple filter, which job is to make sure that only one correspondence is connected with each one of the pattern's features. It is important mechanism, that lowers the false positive detection rate. 

Filters in a cascade can accept aggregated votes or reject them. Cascade consist of two types of filters. Vote data filters make use of data gathered in votes. Graphical filters use additional graphical data extracted from the scene image. Cascade filters comprise of: (1)~vote count thresholding, (2)~adjacency sum thresholding, (3)~scale variance thresholding, (4)~rotation variance thresholding, (5)~feature points binary test, (6)~global normalised luminance cross correlation thresholding. First pass of vote aggregation uses filters: (1), (2), (3) and (4). Second pass of the process uses filters: (3), (4), (5) and (6). 

After successful vote aggregation and analysis, the object's occurrence is assumed. After that, all of the data corresponding to a detected object's area is erased from the vote image and the vote space. Then the next proposition can be analyzed. 

The vote aggregation is the core of the detection system, but the whole framework is much bigger. The detection process for one product is performed in two phases. In phase one each pattern image is resized multiple times, till achieving minimal size. Each derivative pattern is processed as if it was an independent object's pattern. After detection process the occurrence consolidation is performed. It is likely, that the same products will be detected multiple times, as we generated couple of the same patterns but with different size. These detections are merged, and its adjacency sum values are summed. Each occurrence (detection) can be ranked on the basis of the adjacency sum value. The best occurrence is then chosen and a new pattern is extracted straight from the scene image. This new pattern is not resized. In phase two the detection process is performed for a second time for the extracted pattern. Final detections are consolidated and merged with detections from the previous phase. 

After each product has been processed in a way presented earlier, the last consolidation is performed. It is likely, that some of the products in the scene will be detected as a different member of the same brand. Tested implementation doesn't use a multi-stage detection approach. We tested few different wrappings of the same product line to find out the basic detection resolution of the method. If two detections are overlapping, only the one with the best normalized adjacency sum is chosen. Normalization is performed for each pattern independently in regards to amount of visual features detected.

\section{Pattern vs object size}
\noindent The detection efficiency of the presented system depends on the assumption that, if the object exists in the specific area, then one has access to a substantial number of correct votes. We assume also, that the rest of the votes has noise-type distribution over scale, rotation and (x, y) location. 

The system builds a cascade of patterns of different sizes from each one of the base patterns. For each pattern the detection process is performed. This method is a brutal approach, as the computation cost rises with the number of the resized copies. There are multiple benefits though. 

All of the state-of-the-art local features lose repeatability characteristics for images with a different resolution. Matching capability of such features as SIFT or SURF decreases significantly, when the difference in size of the objects in the scene and pattern image are more than 2x. Similar results have been presented in works of Huynh \emph{et al.} \cite{FeaturesEvaluation03}, Khan \emph{et al.} \cite{FeaturesEvaluation02} and Azad \emph{et al.} \cite{FeaturesEvaluation01}. The one reason for this is a fact, that the smaller image will usually has less detected feature points (assuming that both images have similar level of blurring). The average distance difference between two corresponding points in two images is getting bigger with the growth of resolution difference as well. This leads to increased contribution of false matches in the vote space.

To overcome this limitations we decided to create sets of different pattern sizes. This is not only to minimize noise level or boost a matching capability. We can assume, that the number of features in the pattern is not far from the number of features extracted from the object in the scene image (if the object's sizes in the scene and the pattern images are similar). Additionally we can automatically reject all of the matches, which scale difference is over specified range. It's worth mentioning that, if there is a way to determine a real size of the scene frame through some kind of markings on the shelf, the pattern could be resized to the exact size of the object in scene (measured in pixels). This would accelerate processing greatly, yielding extremely low false matching, as the scale difference range could be narrowed down. 

\section{2-phase approach}
\begin{figure}[!htb]
\begin{minipage}[!htb]{\linewidth}
  \centering
  \centerline{\includegraphics[width=4cm]{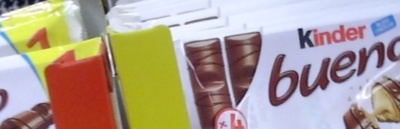}}
  \centerline{False detection taken from phase one.}\medskip
  \label{fig:bad_detection_extracted}
\end{minipage}
  
\begin{minipage}[!htb]{\linewidth}
  \centering
  \centerline{\includegraphics[width=6cm]{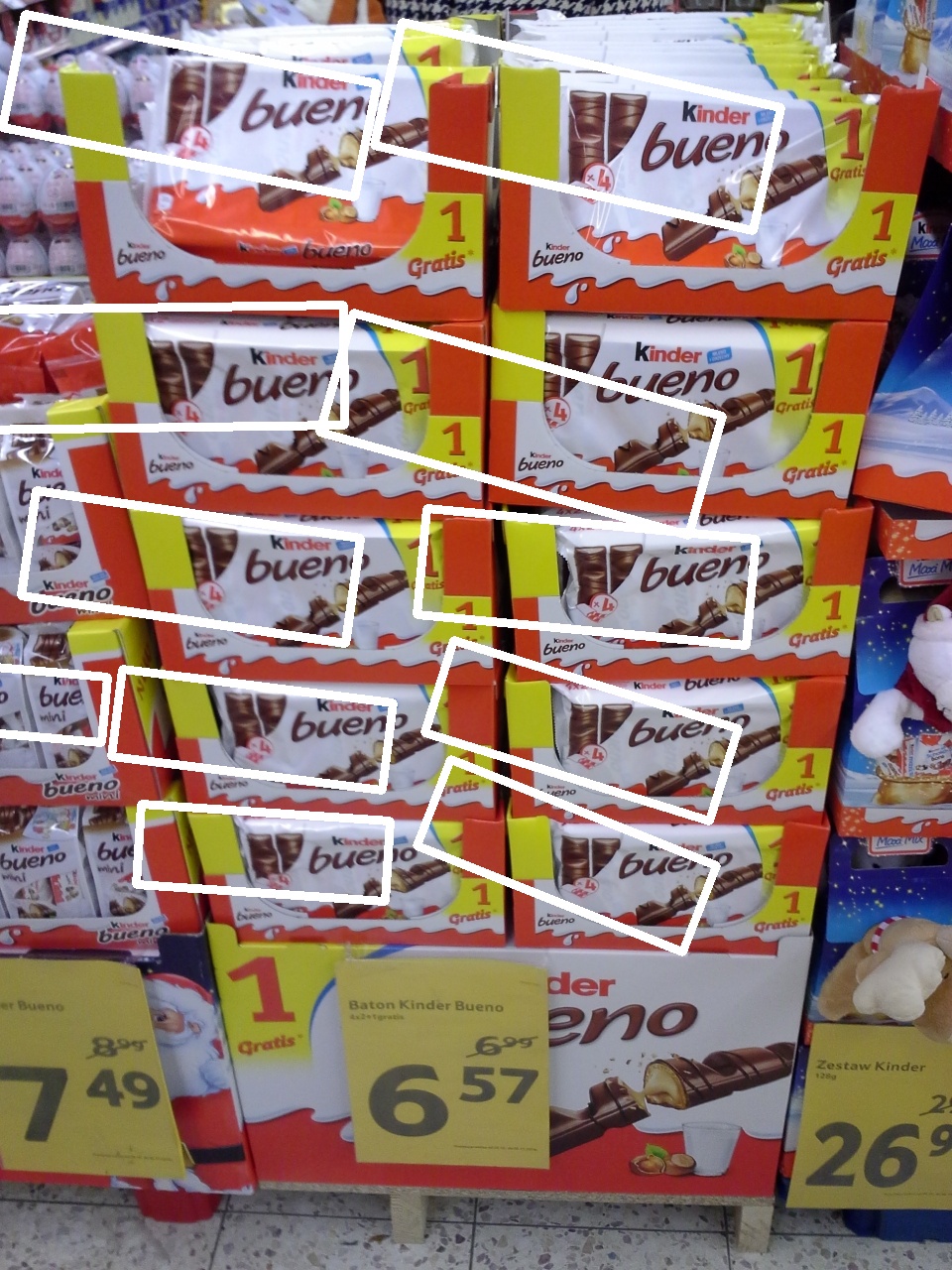}}
  \centerline{Multiple false detections after phase two.}\medskip
  \label{fig:bad_detection_image}
\end{minipage}

\caption{Example of generation of multiple false detections after extracting the false detection from phase one. This example has been achieved by disabling filters in the cascade and using the pattern of object, that is not present in the image. Unfortunately such situation may occur with all filters enabled.}
\label{fig:bad_detection}
\end{figure}

\noindent The 2-phase approach means using the new pattern, extracted from the scene image, for second phase of the detection process. After choosing the best detection in the first phase, we extract the exact area of the detection from the scene image to create a new pattern for second phase of detection. In the second phase there is no resize mechanism, as the new pattern has the exact size of the object in the scene. This mechanism increases computational complexity of the algorithm, but is the only mechanism in the test, that could achieve the highest possible detection rate. All of the modern visual features are susceptible to illumination changes, blurring, perspective warping, noise, bad color representation and many more characteristics of the natural photos. One of the simplest and the most straightforward way to overcome this limitations is to use the image, that is a part of the scene. This operation fits resolution, blur, lighting and noise conditions of the pattern to the conditions of the scene. In most cases, mentioned conditions are uniform for the whole scene. If we have the pattern extracted from the scene, the detection task becomes much easier, reaching even 100\% detection rate for many scenes and objects. Unfortunately, it comes with a problem of false detections in scenes with no objects present. The best detection (a false detection in this case) could be chosen as a new pattern and, as a result, the system could identify the false occurrences in the scene in many other locations. This situation is shown in Figure~\ref{fig:bad_detection}. That's the reason for putting emphasis on lowering the false (negative) detection rate. The first phase does not need to detect multiple objects. It just needs to find one, real occurrence with high certainty. Presented system can be optimized to do such task, lowering the computation cost, as a result of processing only few of the strongest propositions.

\section{Parameters}
\noindent The system can function properly only, if its modules and processes are working jointly with the characteristics of the task. In practise it means many parameter adjustments before the system can be used in practise for broad problem characteristics. This chapter presents some of the main parameters, that must be considered while evaluating effectiveness of the system presented in the paper. 

The first important parameter to determine is a scale factor for resizing the patterns in the first phase of object's detection. We used a scale factor of 2 (for each dimension) for this purpose. Resizing patterns allows narrowing down the scale quotient range in which we accept feature points matches as valid. We found that superimposing the scale acceptance ranges for different pattern sizes does not increase system's effectiveness in a meaningful way. The range for scale quotient has been set to (0.75, 1.5). Theoretically, the narrower the scale acceptance range, the less impact on detection has features' vulnerability for scale difference. Chosen parameters' values have been evaluated with test images and its further adjusting didn't yield any improvement in detections.

The distance threshold for filtering out the matches has been presented earlier in this paper, but it needs a comment. We decided to use half of the distance range, based on intuition and multiple tests, which did not showed any kind of strict correlation or mechanisms, which could lead to calculation of the ideal distance limit. It's mainly because the most reliable success rate metric can be extracted from the detection rate and false (positive) detection chance. Between detections and match filtering there are many other mechanisms that gain or lose its effectiveness with the distance threshold change. Filtering of matches is performed with use of a color filter as well. We reject point correspondences, which has the hue (in HSL color model) difference greater than 45. The filter works only, if the lightness (in HSL) is in range [10, 240] and the biggest difference in RGB channels for each point is over 10. 

During proposition generation we use Good Features To Track algorithm, which has a scanning window parameter. The size of this parameter has big impact on the number of propositions detected and its accuracy. The bigger window can be interpreted as a blurring preprocessing of the vote image.  The detector with too big window can generate inaccurate proposition's locations, which can compromise the aggregation of votes. A small window can generate too many propositions. The size of a scanning window in tests was calculated each time, as:
\begin{equation}
\mbox
{\textit{wSize}}
(pSize) = \left(\floor[\Big]{\frac{pSize}{100}}+1\right)*2+1,
\end{equation}
 where \emph{wSize} is a window size, and \emph{pSize} is a bigger size of the (X, Y) dimensions of the pattern.

During the aggregation process votes are collected in a locality of the proposition. The locality is defined as a window of the same size, as during proposition generation. The Flood Fill algorithm in pass 2 has an aggregation window with the same size as well.

Each one of the filters in the cascade has its parameters. In the vote count thresholding we decided to process only groups of more than 6 votes. The adjacency sum thresholding makes a very similar kind of filter. The adjacency sum threshold is calculated as a number of feature points in the pattern divided by 200. This filter can reject groups of more than 6 votes but with a very weak adjacency values. It's important, that this filter is correlated with the pattern. In the scale variance thresholding we set the scale variance threshold for 60\% of the average value of the scales squared in the aggregated set of votes. The rotation variance is tested in the same way as a scale variance. The difference lays in the calculation method of the rotation variance and average value. The calculation is not straightforward, because of the cyclical character of the rotation metric. 
The feature points binary test compares two binary vectors using Hamming distance. Two binary vectors of the same size are generated for an aggregated vote group - one on the pattern side, and one for the scene side. For each vote pair from the vote group two binary luminance tests are performed. Each test leads to a '1' value for L(p1) > L(p2) and '0' otherwise, where L() is a luminance returning operator, and p1 and p2 are the feature points from the scene (for first binary vector) or from the pattern (for second binary vector). When more than 25\% of the bits are different between the vectors, we reject the vote aggregation. This test is not perfect, as many false detections have differences smaller than 25\%. Nevertheless it can filter out huge amount of false detections, almost not affecting the positive detection rate, as positive aggregation yields very low distances in this test, especially for the phase 2 of the detection. 

The global normalised luminance cross correlation thresholding is the last filter in the cascade. As it can accurately identify almost identical images, it is weak against different frame positioning and lightning conditions. Nevertheless it can filter out some false detections. Because we do not want to reject any positive detections we set the threshold to 0.5 for this algorithm (the cross correlation value's range must be normalised to \emph{(0,1)}). In this filter each color channel is tested independently. The images are resized before the computation to a size of 20x20 pixels. 

\section{Results}

\noindent For experiments we used the same test database as in~\cite{Pse02} for comparison. The image database consists of 120 shelf photos taken in 12 MPx resolution and scaled down to 3 MPx for testing purposes. The pattern group consists of 60 generic patterns of logos and product wrappings. Each shelf photo has been tested with each one of the patterns, conducting 7200 detection processes in total. Each scene contained very few classes of products, so most of the detection processes could generate only false positive detections. Average number of products presented in the scenes was \emph{23.6}. Each pattern has been used with its original size, that was not higher than \emph{700x700} pixels. The biggest ones led to generation of even three resized derivative patterns. Moreover the tests have been performed twice for scenes with the original 12 MPx and with reduced (3 MPx) resolution. The latter can be compared directly with the results of~\cite{Pse02}. The feature points algorithm used for tests was the SIFT feature extractor and detector. 

The testing database is strictly connected with the application of products search. During process of selecting photos for the database the scenes, where the shelf or the face of the products' front were rotated by more than 45 degrees from the photo's scene plane, were ignored. This selection was made manually. 45 degrees criterion gave a big field for error in this process. It is not a crucial problem, as in real application scenes with rotation bigger than 30 degrees can be marked as insufficient, if we want to achieve a detection rate above 90\%. Database contains patterns, which show a whole front of the product as well as only a brand's logo. The patterns' framing have been chosen arbitrarily to test different approaches. Many scenes has very unfavorable lighting conditions and show multiple reflections on the products. Such scenes, connected with an imperfect or a very simple pattern, lead to poor detection rate. On the other hand, visually rich patterns lead to almost perfect detection rate, revealing even products, that are hard to notice for human.

The detection of a brand's logo is associated with a problem of putting the detections to a specific product's group. Some products of the same brand are very similar, with only slight local graphical differences. Presented system can detect a product, even if it is partially occluded. At the same time it can ignore the minor graphical difference and recognize the wrong member of the specific product's line. In real application such detections should be processed further to discriminate different variations of the product. One can use partial patterns with a bag-of-words approach on top of the presented aggregation method to do so. We call it a cascade detection process, where the thorough identification of the product is the result of many sequential algorithms.

\begin{table}[h!]
\begin{center}
\begin{tabular}{|l|c|c|}
\hline
\shortstack{Scene\\size} & Detection Rate & \shortstack{False\\Detection Chance\\(per }\\
\hline
12 MPx & 89.0\% & 0.72\% \\
3 MPx & 84.4\% & 1.63\% \\
\hline
\end{tabular}
\end{center}
\caption{Detection rate and false (positive) detection chance for the tests.}
\label{tab:results}
\end{table}

\begin{table}[h!]
\begin{center}
\begin{tabular}{|l|c|}
\hline
Scene size & Average Number of False Detections\\
\hline
12 MPx  & 3,07\\
3 MPx  & 3,28\\
\hline
\end{tabular}
\end{center}
\caption{Average number of false (positive) detections for a process, when the false (positive) detection occurred.}
\label{tab:results2}
\end{table}

In the Table \ref{tab:results} we showed global results for the tests. We achieved 89\% detection rate for full resolution images. At the 3 MPx resolution we achieved better detection than during tests in ~\cite{Pse02}. Higher resolution yielded lower chance for false detection. The interesting thing is, that in the test with lower resolution we achieved a false detection chance lower than in~\cite{Pse02}, even though the system makes few times more detection processes for different pattern sizes and because of a 2-phase approach. The reason for this result is the dynamic parametrization of the system. This parametrization couldn't prevent the rise in the overall number of false detections, that was more than 3 false detections per image (Table \ref{tab:results2}). This rise is connected with 2-phase approach, that uses the false detection as a new pattern, leading to a multiplication of the false detections.

\section{Conclusions}

\noindent Detection effectiveness of the system lays in three main aspects: proper vote group filtering, good parametrization, well defined pattern. The interesting thing is, that if we decide to use the filters described in this work and adjust the parameters, the pattern choice has the biggest impact on the performance. Size, sharpness levels, noise, lightning conditions - all of this characteristics can lower the detection rate even to 0\% when chosen very unluckily. We found that the parameter-pattern dependencies and pattern extraction from the scene has the biggest impact on the system and should be researched much more. That is definitely a drawback of the one pattern approach, as the learning approaches tend to generalise the descriptor data to fit different application circumstances.

System shows promising results in tests. Simple approach to parameters adjustment and 2-phase processing improved detection ability of the system and is easy to analyze.  System can achieve almost 90\% detection rate with the false detection rate below 1\%, that is acceptable in some real application.

In a future work we will optimize process of proposition acquisition to lower the computational complexity of the system. We are going to evaluate alternative visual features. We will evaluate possibility of using much smaller amount of visual features and a cascade approach to detection process.

\section*{Acknowledgements}

\noindent This work was co-financed by the European Union within the European Regional Development Fund.

%
%


\end{document}